\documentclass{article}
\usepackage{lipsum,spconf,multicol,times, epsfig, graphicx, amsmath, amssymb, floatrow, booktabs, url, multirow, comment,subfigure,mwe}
\usepackage{textcomp}
\graphicspath{ {./figs/} }
\usepackage[T1]{fontenc}

\title{U-SegNet: Fully Convolutional Neural Network based Automated Brain tissue segmentation Tool}
\name{Pulkit Kumar \quad Pravin Nagar \quad Chetan Arora \quad Anubha Gupta}
\address{Indraprastha Institute of Information Technology-Delhi (IIIT-Delhi), Delhi, India.
}

\begin{document}
\maketitle
\begin{abstract}

Automated brain tissue segmentation into white matter (WM), gray matter (GM), and cerebro-spinal fluid (CSF) from magnetic resonance images (MRI) is helpful in the diagnosis of neuro-disorders such as epilepsy, Alzheimer's, multiple sclerosis, etc. However, thin GM structures at the periphery of cortex and smooth transitions on tissue boundaries such as between GM and WM, or WM and CSF pose difficulty in building a reliable segmentation tool. This paper proposes a Fully Convolutional Neural Network (FCN) tool, that is a hybrid of two widely used deep learning segmentation architectures SegNet and U-Net, for improved brain tissue segmentation. We propose a skip connection inspired from U-Net, in the SegNet architetcure, to incorporate fine multiscale information for better tissue boundary identification. We show that the proposed U-SegNet architecture, improves segmentation performance, as measured by average dice ratio, to $89.74\%$ on the widely used IBSR dataset consisting of T-1 weighted MRI volumes of 18 subjects.
\end{abstract}

\section{Introduction}
\label{sec:intro}

Segmentation of brain magnetic resonance imaging (MRI) volume into its basic cytoarchitectural tissue classes is useful for clinicians in the treatment of neurological disorders such as epilepsy, schizohprenia, Alzheimer's, and dementia. Neurologists observe tissue abnormalities of cortical thickening, shrinkage and ventricle expansion for diagnosis and hence, accurate segmentation is crucial for correct diagnosis. Manual segmentation by experts is time consuming, prone to human errors, and impractical for large studies. Hence, development of accurate methods for automated brain tissue segmentation is an active research area.

Automated brain tissue segmentation has primarily three challenges.  Firstly, there are large variations in brain's anatomical structures by phenotype such as age, gender, race, and disease. This leads to difficulty in generalizing one specific segmentation method for all phenotypic categories.  Secondly, challenges are associated with the cytoarchitectural variations such as gyral folds, sulci depths, thin tissue structures, and smooth boundaries between different tissues. This leads to confusion in categorical labeling into distinct tissue classes and is challenging even for a human expert. Lastly, the imaging technology has its own limitations with reference to bias effects of scanner and, signal-to-noise ratio and motion artifacts in the captured MRI images.

Intensity based thresholding \cite{3}, statistical methods \cite{4,greenspan2006constrained,tohka2010brain}, mean-shift \cite{comaniciu2002mean}, adaptive mean-shift \cite{1, mayer2009adaptive} and fuzzy $c$-means clustering \cite{1} are some of the commonly used brain tissue segmentation methods. Of these, intensity thresholding methods perform poorly in low contrast images and at the overlapping boundaries of GM and WM. Statistical methods learn the distribution of the training data and train the parameters accordingly. These may give poor results in the presence of multiplicative bias field \cite{1,mayer2009adaptive}. Mean shift methods estimate mode of the feature vector distribution. However, inappropriate settings of kernel parameters may lead to under or over-segmentation \cite{1}. The problem can be solved by adaptive mean shift methods \cite{1, mayer2009adaptive}. Mahmood et al. \cite{1} used Adaptive Mean Shift (AMS) and fuzzy $c$-means and achieved state-of-the-art performance on the IBSR-18 dataset \cite{IBSR_dataset}.

Currently, with the success of deep learning (DL) methods in different application areas, these techniques are being applied to brain segmentation task as well. Here, researchers have explored both 2D and 3D local neighborhood based methods for the brain tissue segmentation task \cite{8,9}. VoxResNet, Parallel Multi-dimensional Long Short-Term Memory (LSTM), multidimensional gated recurrent unit (GRU) are different variants of deep neural networks that have reported state-of-the-art performance on MICCAI MRBrainS challenge dataset \cite{10, 12, parallel_lstm}. VoxResNet is so far the deepest 3D convolutional network containing $25$ volumetric convolutional layers and $4$ deconvolutional layers. It incorporates 3D information, but given the scarcity of dataset in medical imaging, training a huge network from scratch is a challenging task.

\begin{figure*}[!ht]
	\includegraphics[scale=0.5, trim=0 0 0 10]{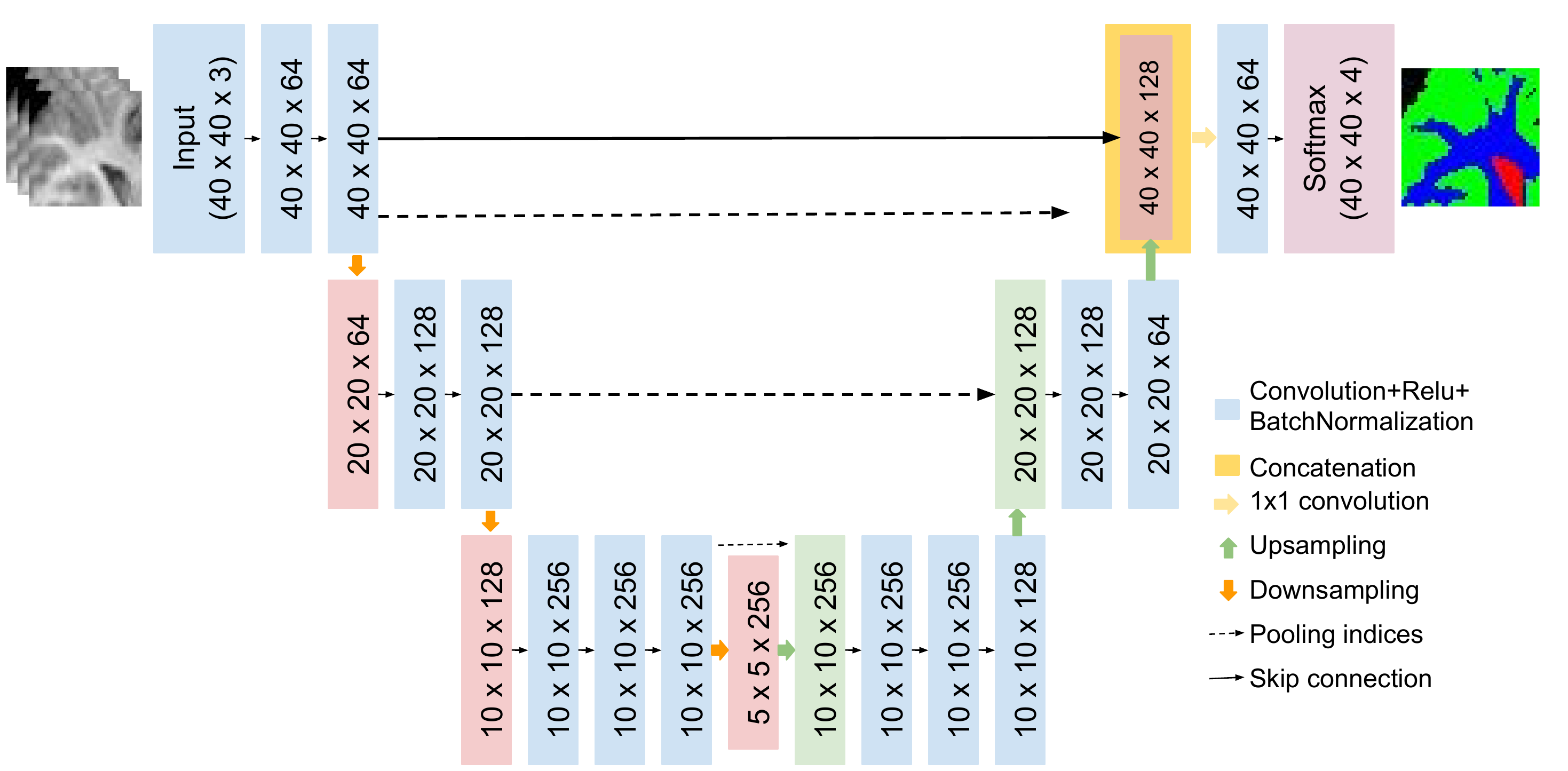}
		\vspace{-1em}
	\caption{We propose a Fully Convolutional Neural Network (FCN), that is a hybrid of two widely used deep learning segmentation architectures SegNet and U-Net, for improved brain tissue segmentation. While the base architecture resembles SegNet, we propose a skip connection inspired from U-Net. These skip connection help the proposed network in capturing fine-grained multiscale information for better tissue boundary identification.}
	\label{fig:archi}
\end{figure*}

The state-of-the-art brain segmentation DL architectures employ 3D models that are computationally heavy and require learning a large number of parameters. This requires a large number of annotated training sample images, while medical imaging data, in general, is limited. This work is motivated towards developing a computationally efficient DL technique that works on limited training data and performs brain image segmentation task with reasonably good performance. In this context, we explore SegNet and U-Net architectures which are stated to require much lesser training data \cite{badrinarayanan2017segnet,u-net}.  SegNet is a well known architecture in computer vision for semantic segmentation \cite{badrinarayanan2017segnet}, but has not been used much so far for the brain MRI segmentation task. It passes pooling indices to the upsampling layers and hence, requires much fewer parameters and is faster to train. U-Net uses multiscale information via skip connections and captures both coarse level and fine level information at the deconvolutional layers \cite{u-net}. However, because of learnable upsampling, U-Net has much larger parameters to learn and is comparatively slower to train than SegNet. On the other hand, SegNet does not capture multiscale information as effectively as the U-Net. We see the complimentary strengths in the two models and explore a combination of the two in this paper. 

\subsection{Contributions}

In this paper, we propose a hybrid architecture of SegNet and U-Net, namely U-SegNet, that captures best of both the models by using SegNet architecture as the base, but with skip connection at selected deconvolutional layer providing multiscale information for better performance. The model has faster convergence because pooling indices are passed to the deconvolutional layers. 

An MRI volume is a 3D data and one can possibly use each volume as a separate sample input. However, these leads to severe reduction in the number of training samples. At the other extreme, researchers have also explored using each slice of the volume as a separate sample. However, this does not exploit the 3D structure in the task. We make a compromise between the two and propose to use 3 slices of the volume as input to segment the middle slice. 

We further observe that giving a full slice (or 3 slices) is detrimental to the performance of the neural network. The large size of the slice increase the number of parameters and makes it harder for the network to effectively learn the parameters. Further, using larger neighborhoods does not necessarily add to the information required for the accurate segmentation. We, thus, propose a segmentation architecture which works on a smaller patch of $40 \times 40 \times 3$ which benefits us in two ways. Firstly, as described above, it reduces the parameters and allows the network to focus on the useful information. Secondly, it allows us to run the model over an image in a sliding window style. This gives multiple output label for each pixel, which we combine by averaging to generate the final label. This is similar to model averaging, shown to be effective my multiple other researchers in their problems, and helps us also in the segmentation task.


\section{Proposed U-SegNet Architecture}

The U-Net architecture is U-shaped model with features of an image learned at different levels through a set of convolutional and max pool layers \cite{u-net}. The feature maps are up-sampled through deconvolutional layers to obtain the segmentation maps at the original image resolution. Also, feature maps of same resolution from down-sampling and up-sampling layers are concatenated in the up-sampling path to incorporate both coarser and finer information.

The SegNet architecture consists of a VGG encoder for the down-sampling path and an inverse VGG for the up-sampling path \cite{badrinarayanan2017segnet}. Unlike U-Net, SegNet does not use deconvolution layers for up-sampling. Instead, feature maps are up-sampled through the pooling indices taken from the down-sampling path at the same resolution. Both SegNet and U-Net architectures use complete images as inputs.

Our proposed U-SegNet architecture is a hybrid of both SegNet and U-Net as shown in Figure \ref{fig:archi}. We posit that the local information is more important than the global information for identifying WM, GM, and CSF. Hence, a patch-based training is adopted on axial slices of size $256 \times 128$. We observed GM structures carefully and noted that for the given resolution IBSR dataset, a patch size of 40 is appropriate to capture sufficient-sized local structures helpful in segmentation. With each patch, equal sized patches from the slice above it and below it are concatenated to add 3D volumetric structural context to the segmentation task.  Overlapping patches shifted by 10 voxels in both directions on the axial slices are captured for training and testing.

We have reduced the depth of the SegNet architecture to handle $40 \times 40 \times 3$ sized input patches. Each convolutional layer uses a 3x3 kernel. Max-pool layers of size $2 \times 2$ and RELU activation functions are used in the architecture. A U-Net type skip connection is introduced only at the upper-most layer as shown in Fig \ref{fig:archi} to incorporate feature maps with fine details. At this layer, a 1x1 convolution layer is used to consolidate coarser and finer information for the segmentation task and also to reduce the number of parameters for the final convolutional layer. The skip connection helps us in incorporating fine information without increasing the parameters as has been done in U-Net. In the end, a softmax layer with 4 outputs is used to implement 4-label classification as background (0), GM (1), WM (2), and CSF (3).

\begin{figure*}[t]
	\begin{center}		\includegraphics[width=1.0\textwidth]{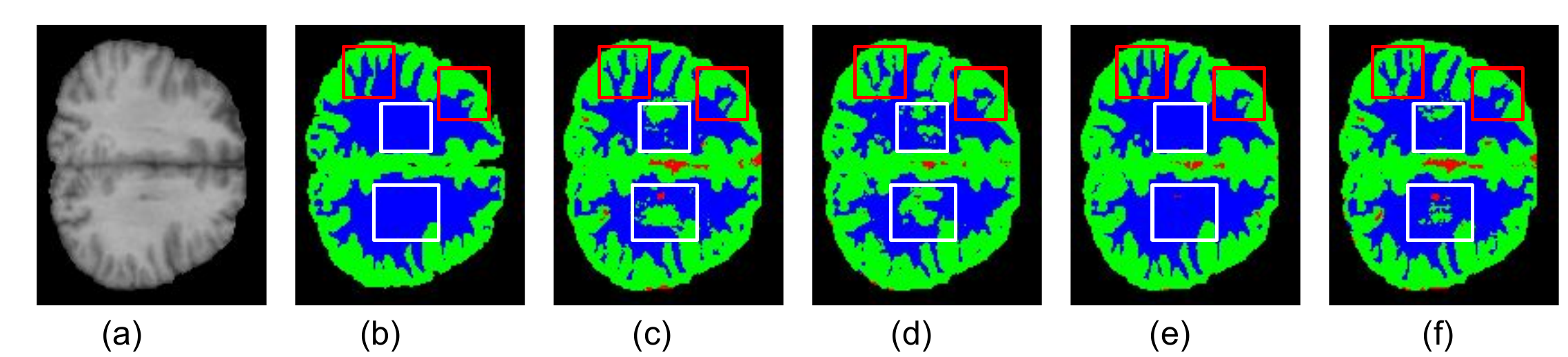}
		\end{center}
	\vspace{-2em}
	\caption{Visualization of (a) Input, (b) Ground Truth labeled image; Black: background, Green: GM, Blue: WM, Red: CSF (c) SegNet, (d) U-Net, (e) Proposed U-SegNet architecture with one skip connection, and (f) Proposed U-SegNet architecture with two skip connections (U-SegNet-2). It is observed that SegNet and U-Net show random patches (white rectangles) and compromise fine details (red rectangles) around the folds generated by gyri and sulci. The proposed architecture captures fine details and solve the random noise problem in the continuous WM seen in U-Net and SegNet. We further observe that adding one more skip connection at the second level in (f) leads to loss of fine information and also fails to handle random noise present in the continuous WM. SegNet, U-Net, and the proposed U-SegNet architecture with two skip connections show that adding skip connections at lower layers leads to performance drop.}
\label{fig:all_brain}
\end{figure*}

\section{Experiments and results}

\subsection{Dataset Description}

We have used IBSR-18 dataset comprising of $18$ T1-weighted MRI volumes of size $256\times128\times256$ of $4$ healthy females and $14$ healthy males with age between $7-71$ years \cite{IBSR_dataset}. These volumes are provided after skull-stripping, normalization and bias field correction. The ground truth is provided with manual segmentation by experts with tissue labels as $0, 1, 2, 3$ for background, CSF, GM, and WM, respectively. Each MRI volume is read, via 256 number of axial brain slices of size $256\times128$ each, in the proposed model.

\subsection{Implementation Details}
Vanilla SegNet was used with weights initialized from a network trained on the CamVid \cite{BrostowSFC:ECCV08,BrostowFC:PRL2008} dataset. Training was done sequentially by fine tuning one layer at a time starting from the last layer with low learning rate of $10^{-6}$. Thus, this new architecture is fine-tuned on the SegNet model for the front-end SegNet layers. The new convolutional layer in the end receiving information via skip connection and layers afterwards are trained from scratch. Stochastic gradient descent (SGD) optimization, batch size of 64, momentum of 0.9, $l_2$ regularisation with parameter $10^{-4}$ were used for a maximum epoch of 700 during the training. Theano with Lasagne was used to train all the models. While the SegNet architecture has 3475396 learnable parameters, U-Net has 3900996, and the proposed U-SegNet has 3483652 parameters. Thus, there is not a substantial increase in the number of parameters compared to the SegNet architecture.

\subsection{Training and Test Data}
We have selected $9$ volumes for training and $9$ for testing. The train-test split comprises all the variation across age and gender. At training time, we selected $6$ volumes for training and $3$ for validation and reported the dice ratio on the test data. For testing, the class of a pixel was decided through majority voting of class labels obtained on overlapping patches. Dice score (DC)  was used as an evaluation metric for all the three tissue classes.
\begin{equation}
\label{eqn}
DC = \frac{2*TP}{2*TP + FP + FN},\nonumber
\end{equation}
where TP, FP, and FN represent the true positives, false positives, and false negatives of the class for which the score is calculated.

\subsection{Results and discussion}

To validate the efficacy of our proposed U-SegNet, we implemented both U-Net and SegNet architectures to benchmark their performance on the IBSR dataset. The quantitative results for each class via average dice score on all 9 test MRI volumes are reported in Table \ref{table:datasets}. Since mean percentage volume of GM, WM, and CSF over the test data are 65.84\%, 32.80\%, and 1.35\%, respectively, weighted dice score (Wt. DC) has also been reported in Table-\ref{table:datasets}.

\renewcommand{\tabcolsep}{0.08cm}
\begin{table}[!ht]
\centering
\begin{tabular}{l c c c c c c}
\toprule[0.2mm]
\textbf{Models}  & \textbf{GM}   & \textbf{WM}   &   \textbf{CSF} & \textbf{Wt. DC} \\
\midrule
Fuzzy $c$-means \cite{1} & 83.11 & \textbf{91.83} & 21.7 & 85.13 \\
\midrule

SegNet \cite{badrinarayanan2017segnet} & 87.36 & 84.15 & 59.04 & 85.92\\
\midrule

U-Net \cite{u-net} & 86.87& 83.58 & 58.36 & 85.40 \\
\midrule

Proposed U-SegNet & \textbf{90.33} & 89.23 & \textbf{66.58} & \textbf{89.64}\\
\midrule

Proposed U-SegNet-2 \\ with two skip connections & 88.17 & 85.95 & 57.81 & 87.03  \\
\bottomrule[0.2mm]
\end{tabular}
\caption{Dice ratio comparison of our method with state-of-the-art approaches. }
\label{table:datasets}
\end{table}

By visualizing the results, we observed that U-Net and SegNet make complimentary errors (Figure \ref{fig:all_brain}). SegNet tends to miss out the finer details especially at the boundary between white matter and gray matter. U-Net on the other hand, because of the skip connections from the lower levels, is able to capture the fine details, say, at the boundaries more accurately than SegNet. However, as shown in Fig \ref{fig:all_brain}(d), U-Net gives errors at places where one class is present in abundance. Also, we observe random noise in the U-Net based segmentation, which we speculate to be because of the confusion created by the deconvolutional layers and skip connections at the lower levels. This can also be possibly attributed to the lack of patch-based training as proposed in our method.

Interestingly, U-SegNet incorporates the good features of both U-Net as well as SegNet. It is observed from Table \ref{table:datasets} that the architecture has the best dice score on GM and CSF and has the second best score (89.23 compared to the best score of 91.83) score on WM. In \cite{1}, apriori spatial tissue probability map generated from brain atlas has been used. This might have resulted in improved segmentation performance over the WM. The proposed method in this work does not utilize any such apriori information. Although WM segmentation performance is slightly inferior (by 1.5\%) compared to \cite{1}, the overall weighted dice ratio has  improved by 4.5\%.

We believe that the $1\times1$ convolutional layer with single skip connection at higher layer consolidates coarser and finer information for the segmentation task. The better capturing of coarser information helps it in reducing random noise in the low frequency or smoother one class region areas observed with U-Net. The better capture of finer information via higher level skip connection (where best finer information is present compared to lower layers) helps with better boundary identification that is the limitation of SegNet.

To further validate the importance of only one skip connection added at the higher layer, we experimented by adding one more skip connection at the second level, called as U-SegNet-2. These results have been reported in the last row of Table \ref{table:datasets}. It is observed that the model starts generating errors of U-Net which could be because of the larger number of parameters added due to multiple skip connections. As a result, the weighted average dice score drops by 2\% compared to the U-SegNet (89.64\% of U-SegNet vis-\`a-vis 87.03\% of U-SegNet-2), although it is still higher from U-Net by 2\% (85.4\% of U-Net).

\section{Conclusion}

Automated brain tissue segmentation is important for disease diagnosis of neurological disorders. In this paper, we have proposed U-SegNet deep learning architecture that is a hybrid of existing SegNet and U-Net architectures. We show that the U-SegNet outperforms state of the art SegNet and U-Net models on the task. Compared to U-Net, U-SegNet has a lesser number of parameters allowing our network to train better. This helps to resolve random noise generated in the U-Net in the proposed architecture. While SegNet tends to miss out on finer details, the proposed model is able to capture these finer details by incorporating the single skip connection in the U-SegNet architecture. We believe that the selective skip connection with 1x1 convolution layer in the upsampling path consolidates both the finer information and the coarser information, improving the segmentation performance. The present work may also find its relevance in other medical imaging applications using deep learning.

\paragraph*{Acknowledgement:}

Chetan Arora has been supported by Infosys Center for Artificial Intelligence and Visvesaraya Young Faculty Research Fellowship by MEITy, Government of India. We acknowledge the support of Prof. Ajay Garg, Department of Neuroimaging and Interventional Neuroradiology, All India Institute of Medical Sciences (AIIMS), New Delhi, India, in helping us understand the data.
    
\bibliographystyle{IEEEbib}
\bibliography{refs}

\begin{thebibliography}{10}

\bibitem{3}
Despotovi{\'c} et~al.,
\newblock ``{MRI} segmentation of the human brain: challenges, methods, and
  applications,''
\newblock {\em Computational and mathematical methods in medicine}, vol. 2015,
  2015.

\bibitem{4}
Marroquin et~al.,
\newblock ``An accurate and efficient bayesian method for automatic
  segmentation of brain {MRI},''
\newblock {\em IEEE transactions on medical imaging}, vol. 21, no. 8, pp.
  934--945, 2002.

\bibitem{greenspan2006constrained}
Greenspan et~al.,
\newblock ``Constrained gaussian mixture model framework for automatic
  segmentation of {MR} brain images,''
\newblock {\em IEEE transactions on medical imaging}, vol. 25, no. 9, pp.
  1233--1245, 2006.

\bibitem{tohka2010brain}
Tohka et~al.,
\newblock ``Brain {MRI} tissue classification based on local markov random
  fields,''
\newblock {\em Magnetic resonance imaging}, vol. 28, no. 4, pp. 557--573, 2010.

\bibitem{comaniciu2002mean}
Dorin Comaniciu and Peter Meer,
\newblock ``Mean shift: A robust approach toward feature space analysis,''
\newblock {\em IEEE Transactions on pattern analysis and machine intelligence},
  vol. 24, no. 5, pp. 603--619, 2002.

\bibitem{1}
Mahmood et~al.,
\newblock ``Automated {MRI} brain tissue segmentation based on mean shift and
  fuzzy $c$-means using a priori tissue probability maps,''
\newblock {\em IRBM}, vol. 36, no. 3, pp. 185 -- 196, 2015.

\bibitem{mayer2009adaptive}
Arnaldo Mayer and Hayit Greenspan,
\newblock ``An adaptive mean-shift framework for {MRI} brain segmentation,''
\newblock {\em IEEE Transactions on Medical Imaging}, vol. 28, no. 8, pp.
  1238--1250, 2009.

\bibitem{IBSR_dataset}
``{IBSR} dataset,'' \url{https://www.nitrc.org/frs/?group_id=48}.

\bibitem{8}
Zhang et~al.,
\newblock ``Deep convolutional neural networks for multi-modality isointense
  infant brain image segmentation,''
\newblock {\em NeuroImage}, vol. 108, pp. 214--224, 2015.

\bibitem{9}
Alexandre de~Br{\'{e}}bisson and Giovanni Montana,
\newblock ``Deep neural networks for anatomical brain segmentation,''
\newblock {\em CoRR}, vol. abs/1502.02445, 2015.

\bibitem{10}
Chen et~al.,
\newblock ``{VoxResNet}: Deep voxel-wise residual networks for volumetric brain
  segmentation,''
\newblock {\em CoRR}, vol. abs/1608.05895, 2016.

\bibitem{12}
Andermatt et~al.,
\newblock ``Multi-dimensional gated recurrent units for automated anatomical
  landmark localization,''
\newblock {\em CoRR}, vol. abs/1708.02766, 2017.

\bibitem{parallel_lstm}
Stollenga et~al.,
\newblock ``Parallel multi-dimensional {LSTM}, with application to fast
  biomedical volumetric image segmentation,''
\newblock in {\em Advances in neural information processing systems}, 2015, pp.
  2998--3006.

\bibitem{badrinarayanan2017segnet}
Badrinarayanan et~al.,
\newblock ``{SegNet}: A deep convolutional encoder-decoder architecture for
  image segmentation,''
\newblock {\em IEEE transactions on pattern analysis and machine intelligence},
  vol. 39, no. 12, pp. 2481--2495, 2017.

\bibitem{u-net}
Ronneberger et~al.,
\newblock ``{U-N}et: Convolutional networks for biomedical image
  segmentation,''
\newblock {\em CoRR}, vol. abs/1505.04597, 2015.

\bibitem{BrostowSFC:ECCV08}
Brostow et~al.,
\newblock ``Segmentation and recognition using structure from motion point
  clouds,''
\newblock in {\em ECCV (1)}, 2008, pp. 44--57.

\bibitem{BrostowFC:PRL2008}
Brostow et~al.,
\newblock ``Semantic object classes in video: A high-definition ground truth
  database,''
\newblock {\em Pattern Recognition Letters}, vol. 30, no. 2, pp. 88--97, 2009.

\end{thebibliography}

\end{document}